\documentclass[letterpaper]{article} % DO NOT CHANGE THIS
\usepackage{aaai23}  % DO NOT CHANGE THIS
\usepackage{times}  % DO NOT CHANGE THIS
\usepackage{helvet}  % DO NOT CHANGE THIS
\usepackage{courier}  % DO NOT CHANGE THIS
\usepackage[hyphens]{url}  % DO NOT CHANGE THIS
\usepackage{graphicx} % DO NOT CHANGE THIS
\usepackage{natbib}  % DO NOT CHANGE THIS AND DO NOT ADD ANY OPTIONS TO IT
\usepackage{caption} % DO NOT CHANGE THIS AND DO NOT ADD ANY OPTIONS TO IT
\usepackage{algorithm}
\usepackage{algorithmic}
\usepackage{xcolor}
\usepackage{colortbl}
\usepackage{multirow}
\usepackage{amsmath,amsfonts}
\usepackage{amssymb}
\usepackage{bbding}

\usepackage{newfloat}
\usepackage{listings}
\DeclareCaptionStyle{ruled}{labelfont=normalfont,labelsep=colon,strut=off} % DO NOT CHANGE THIS
\floatstyle{ruled}
\newfloat{listing}{tb}{lst}{}
\floatname{listing}{Listing}
\title{A Speaker Turn-Aware Multi-Task Adversarial Network \\for Joint User Satisfaction Estimation and Sentiment Analysis}
\author {
     Kaisong Song\textsuperscript{\rm 1,2}, Yangyang Kang\textsuperscript{\rm 1}$^*$, Jiawei Liu\textsuperscript{\rm 3}\thanks{Corresponding authors}, Xurui Li\textsuperscript{\rm 1}, Changlong Sun\textsuperscript{\rm 1}, Xiaozhong Liu\textsuperscript{\rm 4}
}
\affiliations {
    \textsuperscript{\rm 1} Alibaba Group, Hangzhou, China\\
    \textsuperscript{\rm 2} Northeastern University, Shenyang, China\\
    \textsuperscript{\rm 3} Wuhan University, Wuhan, China\\
    \textsuperscript{\rm 4} Worcester Polytechnic Institute, Worcester, USA\\
    kaisong.sks@alibaba-inc.com, yangyang.kangyy@alibaba-inc.com, laujames2017@whu.edu.cn\\ xurui.lxr@alibaba-inc.com, changlong.scl@taobao.com,~xliu14@wpi.edu
}
\usepackage{bibentry}
\begin{document}

\maketitle

\begin{abstract}
User Satisfaction Estimation is an important task and increasingly being applied in goal-oriented dialogue systems to estimate whether the user is satisfied with the service. It is observed that whether the user’s needs are met often triggers various sentiments, which can be pertinent to the successful estimation of user satisfaction, and vice versa. Thus, \textbf{U}ser \textbf{S}atisfaction \textbf{E}stimation (\textbf{USE}) and \textbf{S}entiment \textbf{A}nalysis (\textbf{SA}) should be treated as a joint, collaborative effort, considering the strong connections between the sentiment states of speakers and the user satisfaction. Existing joint learning frameworks mainly unify the two highly pertinent tasks over cascade or shared-bottom implementations, however they fail to distinguish task-specific and common features, which will produce sub-optimal utterance representations for downstream tasks. In this paper, we propose a novel \textbf{S}peaker \textbf{T}urn-Aware \textbf{M}ulti-Task \textbf{A}dversarial \textbf{N}etwork (\textbf{STMAN}) for dialogue-level USE and utterance-level SA. Specifically, we first introduce a multi-task adversarial strategy which trains a task discriminator to make utterance representation more task-specific, and then utilize a speaker-turn aware multi-task interaction strategy to extract the common features which are complementary to each task. Extensive experiments conducted on two real-world service dialogue datasets show that our model outperforms several state-of-the-art methods.
\end{abstract}

\section{Introduction}

The proliferation of complaint and impolite language within service dialogues can create salient social and financial issues, such as marketing harassment and brand reputation damage. Many platforms have built their serving rating systems that rate the suitability and quality of online service. For example, over 77\% of buyers on Taobao\footnote{\url{https://www.taobao.com/}} communicated with sellers before placing an order~\cite{conf/hci/GaoZ11}. User satisfaction estimation (USE) is an important yet challenging task, which has drawn great attention from both industries and research communities~\cite{conf/emnlp/SongBGLZWSLZ19,conf/emnlp/LiuSKHJSLL21,conf/iwsds/HigashinakaDSBF19,conf/www/DengZL0M22}. 

\begin{figure}
  \centering    
  \includegraphics[width=8.3cm]{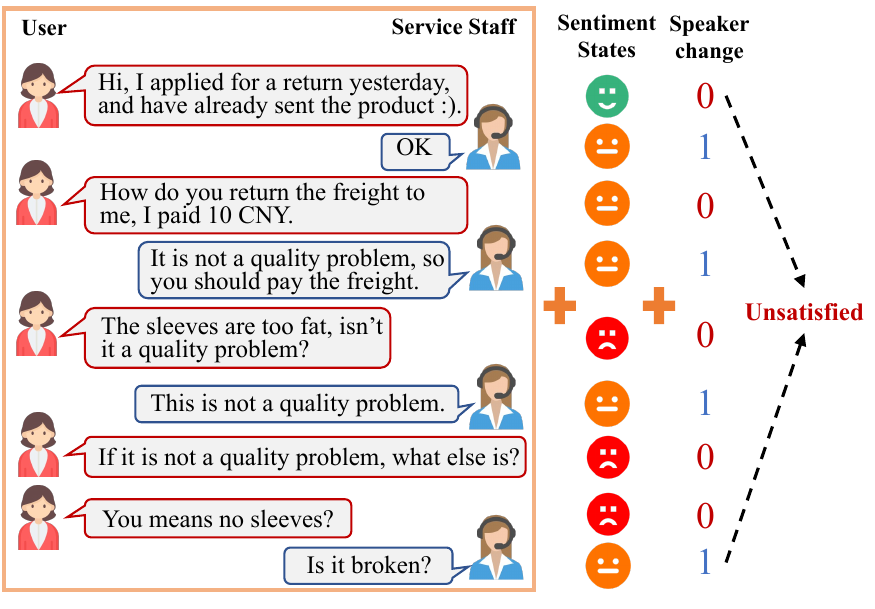}
  \caption{A typical serving rating system and an unsatisfied service dialogue with sentiment and speaker turn change labels. Graphic emoji denotes positive, neutral or negative. Label 1 or 0 denotes whether speaker turn changes or not.}
  \label{fig:examples}
\end{figure}

Intuitively, whether the user is satisfied with a service staff largely depends on the fulfillment of the user's needs. The workflow of a typical service rating system is displayed in Figure~\ref{fig:examples}. The user interacts with the online service staff in the free-form multi-turn dialogue, which is the basic service mode on E-commerce platform. A complete dialogue will be input into the rating system for satisfaction rating. The user expresses positive sentiments in the beginning and interacts with the service staff by using a textual smiley face \textit{``:)''}. As the conversation progresses, sentiment states of the user gradually change from positive to negative, and the user is ultimately very unsatisfied and fails to reach an agreement with the service staff on the return issue. Automatically detecting such low-quality and unsatisfying service is important. For the retail shopkeepers, they can quickly locate such service dialogue and find out the reason to take remedial actions, such as ``\textit{user return visit}" and ``\textit{staff service improvement}". For the platforms, by detecting and analyzing such cases, they can define clear-cut rules, say ``\textit{not fitting well is not a quality issue, and the buyers should pay the freight.}"

User satisfaction often 
relates to many linguistic and statistical properties from content and syntax~\cite{conf/nlpcc/ShanCTX19,journals/jclc/SongWLCCB20,liu2021time}, speaker profiles~\cite{conf/aaai/YaoSLWCCZ20,conf/hci/AbrahamC21} to sentiment~\cite{conf/acit3/GadriMHC21,conf/naacl/SongWCC21}, act~\cite{conf/www/DengZL0M22} and deep semantics~\cite{conf/sigir/Sun0BRRCR21,conf/naacl/KachueeYKL21}, which are regularly considered as key challenges in semantic understanding of dialogues. Different from previous studies that focus on modeling various properties, we find the sentiment states of the speakers provide rich clues for explicitly assisting with user satisfaction estimation.  Without guidance of utterance-level sentiment clues, user satisfaction estimation may become very difficult to understand semantically for long dialogues. Multi-task learning (MTL) allows two or more tasks to be learned jointly, thus sharing information and features between the tasks~\cite{journals/ml/Caruana97}. In this work, we focus on the dialogue-level user satisfaction estimation; hence we refer to it as the \textit{main task}, while the task that is used to provide additional sentiment knowledge, i.e., utterance-level sentiment analysis, is referred to as the \textit{auxiliary task}. Thus, we aim to capture the sentiment feature resulting from a joint learning setup through shared parameters. This will encompass the sentimental content of utterances that is likely to be predictive of user satisfaction. 

Earlier MTL studies on USE often resort to implicit or explicit sentiments, which can not resolve correlated interference between task-private feature and shared feature well. \citet{conf/emnlp/SongBGLZWSLZ19} aggregated the predictive utterance-level latent sentiments as
the estimation of dialogue-level user satisfaction via multiple instance learning~\cite{journals/tacl/AngelidisL18}. However, such cascade implementation will drag performance greatly because sentiment features and satisfaction features will interfere with each other.~\citet{conf/emnlp/LiuSKHJSLL21} considered users' negative sentiments as unsatisfied signals to handover chatting from machine to human. Their shared-bottom implementation alleviates previous problem by simply introducing separate shallow fully-connected layers to derive task-specific utterance features from the same inputs. Despite the effectiveness of incorporating sentiment clues on USE in previous works, there are several issues that remain to be tackled. (i) The performance of task models is influenced by the task-specific input features. Because the shared utterance features can exist in the task private space and the task-specific features creep into the shared space~\cite{conf/aaai/ZhouC0GY21}. (ii) The task models should interact with each other via soft-parameter sharing instead of hard-parameter sharing. Because the later will produce sub-optimal utterance representation for downstream tasks. (iii) Turn-taking modeling across different speakers is necessary, which determines whether sentiment and topics of the consecutive utterances are consistent. For instance, given a \textit{negative} utterance from a user, if the following utterance is from a service staff, then the following sentiment is likely to be \textit{positive} or \textit{neutral}; however, if there is no change in speakers, then the sentiment is probably still \textit{negative}. 

In this paper, we address the joint user satisfaction estimation and sentiment analysis by proposing a novel and extensible Speaker Turn-Aware Multi-Task Adversarial Network (STMAN) model, which contributes in three ways:
\begin{itemize}
    \item We propose a novel multi-task adversarial strategy which uses a task discriminator to differentiate task-specific utterance features explicitly by classifying each utterance into different task categories, i.e, USE and SA.
    \item We propose a novel speaker-turn aware interaction strategy which can extract the common sentiment features of each utterance while each task can still learn its task-specific semantic and speaker-turn change features.
    \item Extensive experiment conducted on two real-world dialogue datasets validate the effectiveness of our approach.
\end{itemize}

\section{Related Work}
\textbf{User Satisfaction Estimation}. User satisfaction, which is related to fulfillment of a specified desire or goal, is essential in evaluating and improving user-centered goal-oriented dialogue systems.~\citet{conf/emnlp/SongBGLZWSLZ19} studied the question-and-answer utterance matching within dialogues and proposed a context assisted multiple instance learning model to predict user satisfaction.~\citet{conf/aaai/YaoSLWCCZ20} conducted a more in-depth exploration of the interactive relationship of multiple rounds of human-computer interaction. They introduced personal historical dialogue to model satisfaction bias. However, both the methods predict satisfaction solely based on content features. Therefore, several efforts have been made on modeling user satisfaction from temporal user behaviors or actions when interacting with the systems~\cite{conf/www/MehrotraLKLH19,conf/emnlp/LiuSKHJSLL21,conf/www/DengZL0M22}.~\citet{conf/www/MehrotraLKLH19} extracted informative and interpretable action sequences (e.g., Click, Pause, Scroll) from user interaction data to predict user satisfaction towards the search systems.~\citet{conf/emnlp/LiuSKHJSLL21} conducted machine-human chatting handoff and service satisfaction analysis jointly.~\citet{conf/www/DengZL0M22} leveraged the sequential dynamics of dialogue acts to facilitate USE in goal-oriented conversational systems via a unified joint learning framework. However, these methods utilize task-specific features in a rough way, thus producing sub-optimal utterance representation for downstream tasks.

\textbf{Dialogue Sentiment Analysis}.  This task assigns a proper sentiment label to each utterance within any dialogue.~\citet{conf/aaai/WangWSLLSZZ20} learned topic enriched utterance representations for utterance-level sentiment classification.~\citet{conf/aaai/QinLCNL21} treated an utterance as a vertex and added an edge between utterances of the same speakers to construct cross-utterances connections; such connections are based on specific speaker roles.~\citet{conf/aaai/MajumderPHMGC19} kept track of the individual party states throughout the conversation and used this information for emotion classification. However, such personalized models have low generalizability because they are severely limited by large amounts of learnable person-related parameters. Previous methods incorporate speaker
information by proposing more complex and specialized models, which inevitably introduce a large
number of parameters to train. Recently,~\citet{conf/emnlp/HeTL021} considered speaker-aware turn changes by concatenating turn embedding and utterance embedding together in a rough way. Here, we utilize sentiment states of speakers to assist with user satisfaction estimation, which is rarely studied before.

\textbf{Multi-Task Learning}. USE and SA can be studied in a joint framework.~\citet{conf/emnlp/BodigutlaTMVP20} proposed to jointly predict turn-level labels and dialogue-level ratings, which used a BiLSTM variant to weight each turn's contribution towards the ratings.~\citet{conf/www/MaGW18} proposed a joint framework that unifies two highly pertinent tasks. Both tasks are trained jointly using weight sharing to extract the common and task-invariant features. However, these frameworks can not be directly adapted to our dialogue scenario well. Firstly, the same dialogue inputs should be treated in a more task-specific way. Secondly, speaker turn change information is ignored, which determines whether sentiment and topics of the consecutive utterances are consistent. Finally, features should be shared in a more controllable way.

\section{Basic Multi-Task Learning Network}
\begin{figure}
  \centering    
\includegraphics[width=8.3cm]{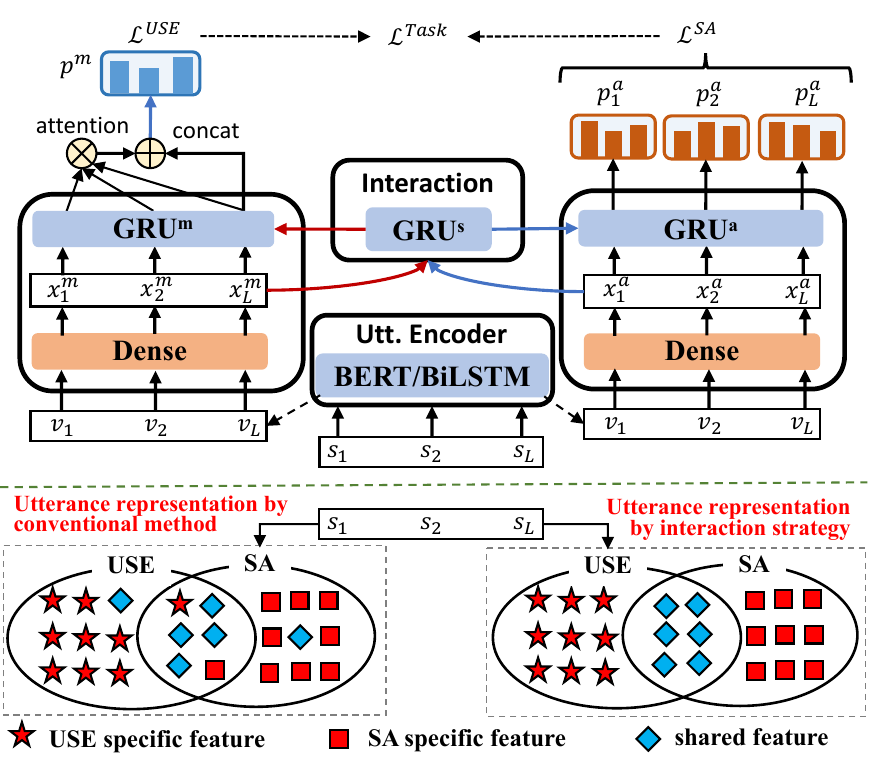}
  \caption{The architecture of Basic model.}
  \label{fig:basic}
\end{figure}

Figure~\ref{fig:basic} shows the overall architecture of our basic model, which consists of four components. This paves way for introducing our fully configured model.

\subsection{Shared Utterance Encoder}
Let any dialogue $\mathcal{D}$ as $[s_1,s_2,...,s_L]$, where $L$ is the number of consisted utterances. The shared utterance encoder $Enc$ is established to map each utterance $s_t\in\mathcal{D}$ to a latent representation:  $\mathbf{v}_t=Enc(s_t)$, where function $Enc$ can be a standard BiLSTM~\cite{journals/tsp/SchusterP97} with attention mechanism~\cite{conf/emnlp/SongBGLZWSLZ19,conf/emnlp/LiuSKHJSLL21} or pre-trained language models, like BERT~\cite{conf/naacl/DevlinCLT19}. Thus, the dialogue $\mathcal{D}$ can be represented as a list of utterance vectors, i.e., $\mathbf{V}=[\mathbf{v}_1,\mathbf{v}_2,...,\mathbf{v}_L]$. Here, we omit the details of the shared utterance encoder due to limited space.

\subsection{Multi-Task Interactive Layer}
The performance of task models
is influenced by the task-specific features. We first apply two dense layers over the shared utterance vectors $\mathbf{V}$ respectively to make them task-specific, which can be formulated as below:
\begin{equation}
\centering   
\mathbf{X}^m=Dense^m(\mathbf{V})\quad,\quad\mathbf{X}^a=Dense^a(\mathbf{V})
\label{eq:dense}
\end{equation}
where vectors $\mathbf{X}^m=\{\mathbf{x}^m_1, \mathbf{x}^m_2,...,\mathbf{x}^m_L\}$ and $\mathbf{X}^a=\{\mathbf{x}^a_1, \mathbf{x}^a_2,...,\mathbf{x}^a_L\}$, superscripts ``$m$'' and ``$a$'' denote the \textit{main} (USE) and \textit{auxiliary} (SA) tasks, respectively. To model the correlations between the separate tasks, we adopt a multi-task interaction strategy which consists of two hidden layers for each task: one is used to extract the common patterns via the shared parameters, and the other is
used to capture task-specific features via the separate parameter sets. Accordingly, each task is assigned a shared GRU layer and a task-specific GRU layer, which hopefully can be used to capture the shared and task-private representations for different tasks. For the \textit{main task}, the USE-specific features $\mathbf{X}^m$ are fed into a soft-parameter sharing standard $\text{GRU}^s$ unit~\cite{conf/emnlp/ChoMGBBSB14}. Meanwhile, we use a variant $\text{GRU}^m$ unit for producing USE-specific utterance representations based on $\mathbf{X}^m$ and outputs of $\text{GRU}^s$. The $\text{GRU}^s$ and $\text{GRU}^m$ units are formulated as below: 
\begin{equation}
\begin{split}
\mathbf{h}^s_t = \text{GRU}^{s}\big(\mathbf{h}^s_{t-1},\mathbf{x}^m_t\big)\\
\mathbf{h}^m_t = \text{GRU}^{m}\big(\mathbf{h}^m_{t-1},\mathbf{h}^s_t,\mathbf{x}^m_t\big) 
\end{split}
\label{eq:main_gru}
\end{equation}
where the hidden output $\mathbf{h}^m_t\in\mathbb{R}^K$ of the $\text{GRU}^{m}$ unit at step $t$ is dependent on the hidden state $\mathbf{h}^s_t\in\mathbb{R}^K$ from the sharing layer, the previous hidden state $\mathbf{h}^m_{t-1}$ from the USE-specific layer, and the current input $\mathbf{x}^m_t$. The hidden state $\mathbf{h}^m_t$ of our $\text{GRU}^m$ can be computed as below:
\begin{equation}
\centering
\begin{split}
r^m_t = \delta\big(\mathbf{W}^m_r\mathbf{x}^m_t+\mathbf{U}^m_r\mathbf{h}^m_{t-1}+\mathbf{U}^{sm}_r\mathbf{h}^s_t\big)\\
z^m_t = \delta\big(\mathbf{W}^m_z\mathbf{x}^m_t+\mathbf{U}^m_z\mathbf{h}^m_{t-1}+\mathbf{U}^{sm}_z\mathbf{h}^s_t\big)\\
\mathbf{\hat{h}}^m_t=\tau\big(\mathbf{W}^m_h\mathbf{x}^m_t+\mathbf{U}^m_\mathbf{h}(\mathbf{h}^m_{t-1} \odot r^m_t)+\mathbf{U}^{sm}_h\mathbf{h}^s_t\big)\\
\mathbf{h}^m_t=\big(1-z^m_t\big) \odot \mathbf{h}^m_{t-1} + z^m_t \odot \mathbf{\hat{h}}^m_t
\end{split}
\label{eq:gru}
\end{equation}
where $\mathbf{U}^{sm}_{*}$ denotes the weight matrix which connects the sharing layer and the task-specific layer, $\mathbf{W}^m_{*}$ and $\mathbf{U}^m_{*}$ are similar to the parameters of GRU$^s$, $\delta(\cdot)$ is the sigmoid function, $\tau(\cdot)$ is the hyperbolic tangent function and $\odot$ is the hadamard product. For the \textit{auxiliary} SA task, we also have the similar operation via the $\text{GRU}^s$ from the sharing layer and a $\text{GRU}^a$ from a SA-specific layer as below:
\begin{equation}
\begin{split}
\mathbf{h}^s_t = \text{GRU}^{s}(\mathbf{h}^s_{t-1},\mathbf{x}^a_t)\\
\mathbf{h}^a_t = \text{GRU}^{a}(\mathbf{h}^a_{t-1},\mathbf{h}^s_t,\mathbf{x}^a_t) 
\end{split}
\label{eq:aux_gru}
\end{equation}

\subsection{Decoder for User Satisfaction Estimation} 
User satisfaction is influenced by role and position information~\cite{conf/emnlp/SongBGLZWSLZ19,conf/emnlp/LiuSKHJSLL21}. Intuitively, user satisfaction is more related to the utterances of users rather than that of service staffs who usually express positive and neutral emotions, and utterances at the end of a dialogue tend to reflect the user's final attitude. Thus, we first adopt a role-selected mask mechanism to reserve the hidden states of all the user utterances and then produce a user-specific vector. Besides, we measure the importance of each user utterance through a scoring function as below:
\begin{equation}
\begin{split}
\mathbf{u}_t = \tanh(\mathbf{W}_u \mathbf{h}^m_t+\mathbf{b}_u)\\
\alpha_t=\frac{exp\Big(mask_r\big(\mathbf{u}^{\text{T}}_t\mathbf{U}_u\big)\Big)}{\sum^{L}_{k=1}exp\Big(mask_r\big(\mathbf{u}^{\text{T}}_k \mathbf{U}_u\big)\Big)}
\end{split}
\label{eq: use_decoder}
\end{equation}
where $\mathbf{W}_u\in\mathbb{R}^{H\times K}$, $\mathbf{b}_u\in\mathbb{R}^H$ and $\mathbf{U}_u\in\mathbb{R}^H$ are trainable model parameters, function $mask_r(x)$ returns value $x$ if speaker role $r=user$, otherwise $-\infty$. Therefore, we obtain the user-specific vector via the weighted sum of the hidden states of all the user utterances. In order to balance roles and positions, the final dialogue representation can be obtained by concatenating the user-specific vector and the last hidden output of $\text{GRU}^m$ as follow:
\begin{equation}
\mathbf{o}^m = \Big[\big(\sum_{t\in[1,L]} \alpha_t \mathbf{h}^m_t \big), \mathbf{h}^m_L\Big]
\end{equation}
where $[,]$ denotes the vector concatenation, the vector $\mathbf{o}^m\in\mathbb{R}^{2K}$ will be input into a linear layer and then a softmax layer for obtaining user satisfaction estimation as below:
\begin{equation}
\mathbf{p}^m = \text{softmax}(\mathbf{W}^m\mathbf{o}^m+\mathbf{b}^m)
\end{equation}
where $\mathbf{W}^m\in\mathbb{R}^{|\Omega|\times 2K}$ and $\mathbf{b}^m\in\mathbb{R}^{|\Omega|}$ are learnable model parameters, and user satisfaction labels $\Omega=\{unsatisfied, met, well~ satisfied\}$.

\subsection{Decoder for Sentiment Analysis}
For the SA task, each utterance vector $\mathbf{h}^a_t\in\mathbb{R}^K$ from GRU$^a$ unit will be first input into a linear layer and then a softmax layer for calculating the sentiment probability $\mathbf{p}^a_t\in\mathbb{R}^{\mathcal{|G|}}$:
\begin{equation}
\mathbf{p}^a_t = \text{softmax}(\mathbf{W}^a\mathbf{h}^a_t + \mathbf{b}^a)
\end{equation}
where $\mathbf{W}^a\in\mathbb{R}^{|\mathcal{G}| \times K}$, $\mathbf{b}^a\in\mathbb{R}^{|\mathcal{G}|}$ are learnable parameters, $\mathcal{G}=\{negative, neutral, positive\}$ are sentiment classes.

\subsection{The Training Procedure} 
For the USE and SA tasks, we have the single-task loss functions $\mathcal{L}^{USE}$ and $\mathcal{L}^{SA}$, respectively, in Formula~\ref{eq:loss_task}. To train our multi-task basic model, we sum the above loss functions up as the final multi-task loss function:
\begin{equation}
\mathcal{L}^{Task} = \underbrace{-\sum_{l\in \Omega} \mathbf{g}^m_l \log(\mathbf{p}^m_l)}_{\mathcal{L}^{USE} ~in~main~task} \underbrace{-\frac{1}{L}\sum_{s\in \mathcal{D}}\sum_{l \in \mathcal{G}} \mathbf{g}^a_{sl} \log(\mathbf{p}^a_{sl})}_{\mathcal{L}^{SA}~in~auxiliary~task}
\label{eq:loss_task}
\end{equation}
where $\mathbf{g}^m_l$ and $\mathbf{p}^m_l$ are respectively the ground truth and the predicted probability corresponding to the $l$-th satisfaction class. $\mathbf{g}^a_{sl}$ and $\mathbf{p}^a_{sl}$ are respectively the ground truth and the predicted probability corresponding to the $l$-th sentiment class for each utterance. We use back propagation to calculate the gradients of all the trainable parameters, and update them with momentum optimizer~\cite{journals/nn/Qian99}.

\section{STMAN: Speaker Turn-Aware Multi-Task Adversarial Network}
\begin{figure}
  \centering    
\includegraphics[width=8.3cm]{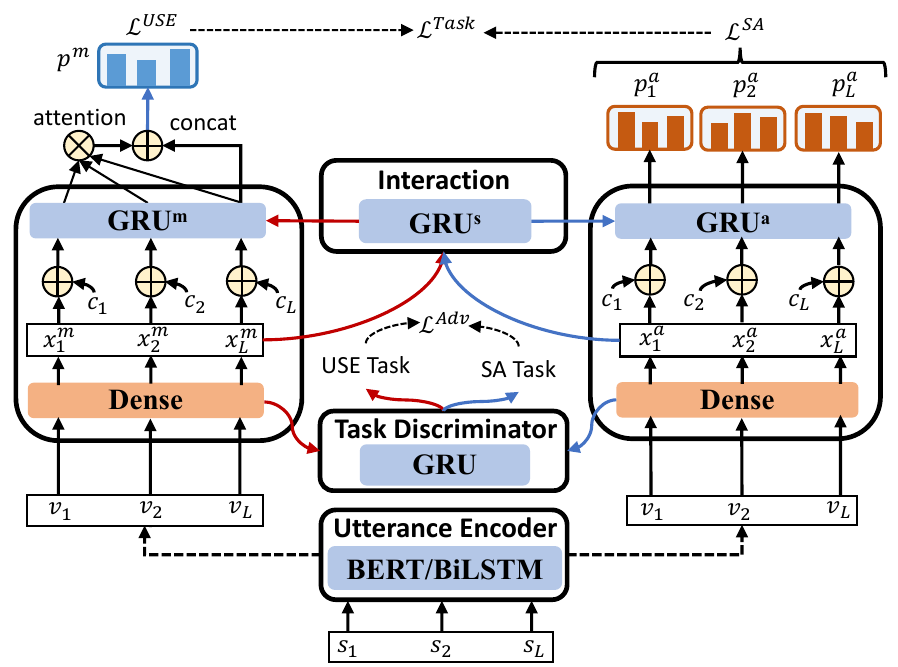}
  \caption{The architecture of our enhanced model: STMAN.}
  \label{fig:STMAN}
\end{figure}

In conventional multi-task learning, the performance of task models is influenced by the task-specific features. Intuitively, the dense layers in \textit{Basic} model can not produce task-specific features well without supervision signals. Thus, we introduce additional \textit{Multi-Task Adversarial Layer} which uses a \textit{Task Discriminator} to differentiate task-specific input features explicitly. Besides, we consider speak-turn change information and introduce additional \textit{Speaker Turn-Aware MT Interactive Layer}. Finally, we build a novel and extensible \textbf{S}peaker \textbf{T}urn-Aware \textbf{M}ulti-Task \textbf{A}dversarial \textbf{N}etwork (\textbf{STMAN}) and display its architecture in Figure~\ref{fig:STMAN}.

\subsection{ST-Aware MT Interactive Layer}
Different from speaker role-based methods~\cite{conf/aaai/QinCLN020}, our method focuses on speaker turns and thus is still useful when speakers are not associated with specific roles. Specifically, we relabel the speakers and flip the speaker label (from 0
to 1 and vice versa) when there is speaker turn change. For the example in Figure~\ref{fig:examples}, if the original speaker sequence is $<$U,S,U,S,U,S,U,U,S$>$\footnote{U denotes a user and S represents a service staff.}, the relabeling sequence is $<$0,1,0,1,0,1,0,0,1$>$, which can be represented by the two introduced speaker turn embeddings in further. Given the utterance representation $\mathbf{x}^*_t$ and its binary
speaker turn label $r_{s_t}\in\{0, 1\}$, we have role enhanced utterance representation $[\mathbf{x}^*_t, \mathbf{c}_t]$, where $\mathbf{c}_t=Emb(r_{s_t})$ denotes any speaker turn embedding and $\mathbf{c}_t\in\mathbb{R}^Z$. Finally, the task-specific GRU units in Formulas~\ref{eq:main_gru} and~\ref{eq:aux_gru} can be reformulated as below: 
\begin{equation}
\centering
\begin{split}
\mathbf{h}^m_t = \text{GRU}^{m}\big(\mathbf{h}^m_{t-1},\mathbf{h}^s_t,[\mathbf{x}^m_t, \mathbf{c}_t]\big) \\
\mathbf{h}^a_t = \text{GRU}^{a}\big(\mathbf{h}^a_{t-1},\mathbf{h}^s_t,[\mathbf{x}^a_t, \mathbf{c}_t]\big)
\end{split}
\label{eq:gru}
\end{equation}
Obviously, $\mathbf{h}^m_a$ and $\mathbf{h}^a_t$ contain speaker turn change information, which are helpful for downstream USE and SA tasks. 

\subsection{MT Adversarial Layer}
Without any supervised signals, the two dense layers (see Formula~\ref{eq:dense}) can not differentiate task-specific features accurately, thus the shared features can exist in the task private space and the task-specific features creep into the shared space. Intuitively, adversarial learning can be further applied to multi-task learning, thus task-specific features $\mathbf{X}^m$ and $\mathbf{X}^a$ will be more distinguishable. Here, we introduce a \textit{Task Discriminator} (TD). Specifically, we exploit the min-max game between the shared utterance encoder and the task-specific dense layer with the task discriminator~\cite{conf/nips/GoodfellowPMXWOCB14}. TD is composed of a GRU layer stacked with a softmax layer to estimate what kind of tasks the utterance representations come from, i.e., it maps the task-specific features into a probability distribution $\mathbf{p}^k\in\mathbb{R}^{|\mathcal{K}|}$, where task type $k\in\mathcal{K}$ and $\mathcal{K}=\{USE, SA\}$. Thus, we use a task adversarial loss $\mathcal{L}^{Adv}$ to limit the task-specific features into their private feature space, which is defined as follow:
\begin{equation}
\mathcal{L}^{Adv} = \min\limits_{\theta_{Enc}}\Big( \;\max\limits_{\theta_{TD},~\theta_{k}} \sum_{s\in \mathcal{D}} \sum_{l\in\mathcal{K}}\mathbf{g}^{k}_{sl} \log\big(\mathbf{p}^{k}_{sl}\big)\Big)
\label{eq:loss_adv}
\end{equation}
where $\mathbf{g}^{k}_{sl}$ and $\mathbf{p}^{k}_{sl}$ are respectively the ground-truth and the predicted probability corresponding to the $l$-th task class for each utterance in the dialogue $\mathcal{D}$. $\theta_{Enc}$, $\theta_{TD}$ and $\{\theta_{k}\}$ are the learnable parameters of \textit{Shared Utterance Encoder}, \textit{Task Discriminator} and respective \textit{Decoders}. And the whole training procedure is shown in algorithm~\ref{alg:algorithm}.
\begin{algorithm}[tb]
\small
\caption{Multi-Task Adversarial Learning}
\label{alg:algorithm}
\textbf{Input}: Training set $\{\mathcal{D}\}$; \\
\textbf{Initialized Parameters}: shared utterance encoder $\theta_{Enc}$, task private decoders $\{\theta_{k}|k\in\mathcal{K}\}$ and task discriminator $\theta_{TD}$;\par
\begin{algorithmic}[1] %[1] enables 
\FOR{mini-batches in $\{\mathcal{D}\}$}
\STATE Minimize $\mathcal{L}^{Adv}$ in Eq.~\ref{eq:loss_adv} to update $\theta_{Enc}$;
\STATE Maximize $\mathcal{L}^{Adv}$ in Eq.~\ref{eq:loss_adv} to update $\theta_{TD}$ and $\{\theta_{k}\}$;
\ENDFOR\\
\FOR{mini-batches in $\{\mathcal{D}\}$}
\STATE Minimize $\mathcal{L}^{Task}$ in Eq.~\ref{eq:loss_task} to update all the trainable parameters except for $\theta_{TD}$;
\ENDFOR
\end{algorithmic}
\textbf{Output}: The well trained $\theta_{Enc}$, $\theta_{TD}$ and $\{\theta_{k}\}$
\end{algorithm}

\section{Experiments and Results}
In this section, we conduct extensive experiments on two publicly available service dialogue datasets to evaluate the effectiveness of our approach. \textit{Clothes} and \textit{Makeup} datasets are collected by~\citet{conf/emnlp/SongBGLZWSLZ19} from E-commerce platform \url{taobao.com} and split into training/development/testing sets with splits 8/1/1. The \textit{Clothes} is a larger imbalanced dataset and the \textit{Makeup} is a smaller balanced dataset. The statistics of datasets are given in Table~\ref{tab:dataset}.

\begin{table}
\centering
\begin{tabular}{|l|c|c|}
\hline
% \rowcolor{gray!20}
\textbf{Statistics items} & \textbf{Clothes} & \textbf{Makeup}\\\hline
\# Dialogues & 10,000 & 3,540\\
\# US (unsatisfied) & 2,302 & 1,180\\
\# MT (met) & 6,399 & 1,180\\ 
\# WS (well satisfied) & 1,299 & 1,180\\
\hline\hline
\# Utterances & 123,242 & 46,255\\
\# NG (negative)  & 12,619 & 6,130\\
\# NE (neutral)  & 97,380 & 33,158\\
\# PO (positive) & 13,243 & 6,976\\\hline
\end{tabular}
\caption{The statistics of the two benchmark datasets.}
\label{tab:dataset}
\end{table}

\subsection{Experimental Settings}
We encode utterances with BiLSTM as that used in works \cite{conf/emnlp/SongBGLZWSLZ19,conf/emnlp/LiuSKHJSLL21} because BERT-based encoder is a sub-optimal choice considering different types of pre-training corpus. Trainable model parameters are initialized by sampling values from a uniform distribution $\mathcal{U}(-0.01, 0.01)$. The hyper-parameters are tuned to the best on the development set. The size of hidden states $K$ is 100, speaker turn embedding size $Z$ is 100, $H$ is 50, the dropout rate is 0.2, the learning rate is 0.1, the learning rate decay is 0.8, the batch size is 32 and the number of epochs is 30. We use Macro F1 and Accuracy as the evaluation metrics. Our programs are implemented by tensorflow\footnote{\url{https://tensorflow.google.cn/}} and run on a server configured with a Tesla V100 GPU, 2 CPU and 32G memory. All the resources are publicly available\footnote{\url{https://sites.google.com/view/ssa-sa/}}.

\subsection{Comparative Study}
In Table~\ref{tab:methods_cmp_clothes} and Table~\ref{tab:methods_cmp_makeup}, we compare our approach with several state-of-the-art USE single-task models (1-4) and the multi-task models (5-8) over the two dialogue datasets.

\begin{itemize}
  \item \textbf{BERT+LSTM} encodes each utterance with BERT-based encoder and uses the last hidden state of LSTM for satisfaction estimation~\cite{conf/emnlp/LiuSKHJSLL21}. 

  \item \textbf{MILNET} represents a Multiple Instance Learning Network for document-level and sentence-level sentiment analysis~\cite{journals/tacl/AngelidisL18}. In~\citet{conf/emnlp/SongBGLZWSLZ19}, USE is considered as a special SA task by treating dialogue as document and utterance as sentence.
  
  \item \textbf{CAMIL} is a Context-Assisted Multiple Instance Learning model which predicts the utterance-level sentiments of all the user utterances and then aggregates the sentiments into user satisfaction estimation~\cite{conf/emnlp/SongBGLZWSLZ19}. 
  \item \textbf{SLUP} denotes a session-level user satisfaction estimation method trained on a sequence of question-answer pairs. To adapt to this method, we produce Q-A pairs by considering consecutive user utterances as questions and consecutive staff utterances as answers~\cite{conf/aaai/YaoSLWCCZ20}.

  \item \textbf{MT-ES} is an Enhanced Shared-layer Multi-Task architecture with two hidden layers: one is used to extract the common patterns via the shared parameters, and the other is used to capture task-specific features via the separate parameter set~\cite{conf/www/MaGW18}. 

  \item \textbf{MT-US} is a simplified version of MT-ES with the uniform shared-layer~\cite{conf/coling/CerisaraJOL18}. This belongs to a kind of shared-bottom multi-task model.

  \item \textbf{Meta-LSTM} is a shared meta-network to capture the meta-knowledge of semantic composition, which conducts text classification and sequence tagging~\cite{conf/aaai/ChenQLH18}. We input the utterance vectors directly, and output dialogue-level USE label and utterance-level SA labels.

  \item \textbf{RSSN} conducts machine-human chatting handoff (i.e., \textit{transferable} or \textit{normal}) and service satisfaction analysis jointly. For the three-class SA task, RSSN is tailored and tuned best on the development sets~\cite{conf/emnlp/LiuSKHJSLL21}.

  %\item \textbf{USDA} estimates user satisfaction and recognizes dialogue act jointly with a unified model. We consider sentiment labels as special act labels~\cite{conf/www/DengZL0M22}.
\end{itemize}

Note that MILNET and CAMIL output utterances' sentiment predictions implicitly derived from satisfaction labels.

\begin{table}
  \centering
  \setlength\tabcolsep{1pt}
  \begin{tabular}{|l|c|c|c|c|}
  \hline
  % \rowcolor{gray!20}
  & \multicolumn{2}{c|}{\textbf{USE} (main)} &\multicolumn{2}{c|}{\textbf{SA} (auxiliary)}\\
    \cline{2-5}
    \multirow{-2}{*}{\textbf{Methods}} & \scriptsize{Macro F1} & \scriptsize{Accuracy} & \scriptsize{Macro F1} & \scriptsize{Accuracy}\\
    \hline
    BERT+LSTM~\scriptsize{\cite{conf/emnlp/LiuSKHJSLL21}} &67.90 &75.50 &- &-\\   
    MILNET~\scriptsize{\cite{conf/emnlp/SongBGLZWSLZ19}} & 63.81  & 75.34  & 55.32 & 71.33 \\
    SLUP~\scriptsize{\cite{conf/aaai/YaoSLWCCZ20}} &66.56 &75.70 &- &-\\
    CAMIL~\scriptsize{\cite{conf/emnlp/SongBGLZWSLZ19}}  & 70.40  & 78.31 & 64.44 & 82.43  \\
    \hline
    MT-US~\scriptsize{\cite{conf/coling/CerisaraJOL18}} & 68.25  & 74.60  & 80.10 & 92.65 \\
    MT-ES~\scriptsize{\cite{conf/aaai/MajumderPHMGC19}} & 68.04  & 74.20  & 80.19  & 90.78\\
    Meta-LSTM~\scriptsize{\cite{conf/aaai/ChenQLH18}} & 69.05  & 77.40 & 80.08 & 92.56\\
    %USDA~\scriptsize{\cite{conf/www/DengZL0M22}} &69.03 &76.40 &76.14 &92.31\\
    RSSN~\scriptsize{\cite{conf/emnlp/LiuSKHJSLL21}} & 69.51 & 77.20 & 80.29 & 91.15 \\
    \hline
    \textbf{STMAN}~\scriptsize{(See Figure~\ref{fig:STMAN})} &\textbf{71.11} &\textbf{78.60} &\textbf{81.90} &\textbf{93.05}\\
  \hline
  \end{tabular}
  \caption{Comparison among different models on \textit{Clothes}.}
  \label{tab:methods_cmp_clothes}
\end{table}

\begin{table}
  \centering
  \setlength\tabcolsep{1pt}
  \begin{tabular}{|l|c|c|c|c|}
  \hline
  % \rowcolor{gray!20}
  & \multicolumn{2}{c|}{\textbf{USE} (main)} &\multicolumn{2}{c|}{\textbf{SA} (auxiliary)}\\
    \cline{2-5}
    \multirow{-2}{*}{\textbf{Methods}} & \scriptsize{Macro F1} & \scriptsize{Accuracy} & \scriptsize{Macro F1} & \scriptsize{Accuracy}\\
    \hline
    BERT+LSTM~\scriptsize{\cite{conf/emnlp/LiuSKHJSLL21}} &74.80 &74.80 &- &-\\ 
    MILNET~\scriptsize{\cite{conf/emnlp/SongBGLZWSLZ19}} & 75.30 & 75.19 & 41.71  & 41.09\\
    SLUP~\scriptsize{\cite{conf/aaai/YaoSLWCCZ20}} &76.31 &76.28 &- &-\\
    CAMIL~\scriptsize{\cite{conf/emnlp/SongBGLZWSLZ19}} & 78.60 & 78.59 & 59.54  & 64.73  \\
    \hline
    MT-US~\scriptsize{\cite{conf/coling/CerisaraJOL18}} & 76.37 & 75.98  & 82.12 & 90.77 \\
    MT-ES~\scriptsize{\cite{conf/aaai/MajumderPHMGC19}} &76.06  & 76.83  & 80.51  & 89.96\\
    Meta-LSTM~\scriptsize{\cite{conf/aaai/ChenQLH18}}  &76.65 &76.55 &80.13 &90.48\\
    %USDA~\scriptsize{\cite{conf/www/DengZL0M22}} &77.11 &77.10 &\textbf{85.23} &91.66\\
    RSSN~\scriptsize{\cite{conf/emnlp/LiuSKHJSLL21}} &79.18 & 79.17 & 83.36 & 90.42\\\hline
    \textbf{STMAN}~\scriptsize{(See Figure~\ref{fig:STMAN})} & \textbf{80.11} & \textbf{80.22} & \textbf{84.18} & \textbf{91.54}\\
  \hline
  \end{tabular}
  \caption{Comparison among different models on \textit{Makeup}.}
  \label{tab:methods_cmp_makeup}
\end{table}

\textbf{Results and Analysis:} From Table~\ref{tab:methods_cmp_clothes} and Table~\ref{tab:methods_cmp_makeup}, we can find MILNET performs worse than SLUP because it ignores dialogue structure information. SLUP achieves substantial improvements by modeling dialogue interactions precisely, but it still ignores the usage of sentiment knowledge. CAMIL performs better than BERT+LSTM by aggregating sentiment predictions directly into satisfaction polarity, however its cascade structure may introduce noisy features. MT-US, MT-ES and Meta-LSTM are typical MT models, which perform not as well as RSSN because they are general frameworks and ignore dialogue structure which influences model performance greatly. Compared with all these baselines, our STMAN achieves the best results in both the tasks because it obtains better task-specific features by MT adversarial strategy and models utterance interaction well by ST-aware MT interaction strategy.

\subsection{Ablation Study}
Different model configurations can largely affect the model performance. In Table~\ref{tab:ab_study}, we implement several model variants for ablation study by removing (``-") and adding (``+") different model components. \textbf{Basic} is the basic method ignoring all the strategies (see Figure~\ref{fig:basic}). \textbf{Basic-Mask} removes mask operation in Formula~\ref{eq: use_decoder} and reserves all the speakers' utterances. \textbf{Basic-Aux} removes auxiliary SA task. \textbf{Basic+TD} enhances \textbf{Basic} by considering task discriminator. \textbf{Basic+ST} enhances \textbf{Basic} by considering speaker turn changes. \textbf{Basic+TD+ST} is our fully configured model.

\begin{table}
  \centering
  \setlength\tabcolsep{1pt}
  \begin{tabular}{|l|c|c|c|c|}
  \hline
  % \rowcolor{gray!20}
  & \multicolumn{2}{c|}{\textbf{USE} (main)} &\multicolumn{2}{c|}{\textbf{SA} (auxiliary)}\\
    \cline{2-5}
    \multirow{-2}{*}{{\textbf{Methods on Clothes}}} & \scriptsize{Macro F1} & \scriptsize{Accuracy} & \scriptsize{Macro F1} & \scriptsize{Accuracy}\\
    \hline
    Basic~\scriptsize{(See Figure~\ref{fig:basic})}  &67.67   &74.40  &81.05  &92.36  \\
    Basic-Mask &67.29 &73.70 &79.70 &92.26\\
    Basic-Aux &67.03 &73.31 & - & -\\
 Basic+TD  &68.92  &75.80   &81.52   &92.64\\
 Basic+ST  &69.53   &76.70   &81.54   &92.48 \\
 Basic+TD+ST~\scriptsize{(STMAN)}  &\textbf{71.11} &\textbf{78.60} &\textbf{81.90} &\textbf{93.05}\\
  \hline
  \end{tabular}
  \begin{tabular}{|l|c|c|c|c|}
  \hline
  % \rowcolor{gray!20}
  & \multicolumn{2}{c|}{\textbf{USE} (main)} &\multicolumn{2}{c|}{\textbf{SA} (auxiliary)}\\
    \cline{2-5}
    \multirow{-2}{*}{{\textbf{Methods on Makeup}}} & \scriptsize{Macro F1} & \scriptsize{Accuracy} & \scriptsize{Macro F1} & \scriptsize{Accuracy}\\
    \hline
    Basic~\scriptsize{(See Figure~\ref{fig:basic})}   & 78.69  & 78.71 &83.01   &88.96 \\
     Basic-Mask  &77.12   &77.11  & 82.41 & 88.62 \\
     Basic-Aux &74.64 &74.57 & - & -\\
  Basic+TD  & 79.05  & 79.09  & 83.03  & 91.30\\
  Basic+ST & 79.03 & 79.05 & 83.30 & 91.22\\
  Basic+TD+ST~\scriptsize{(STMAN)}  & \textbf{80.11} & \textbf{80.22} & \textbf{84.18} & \textbf{91.54}\\
  \hline
  \end{tabular}
  \caption{Comparison among different model configurations.}
\label{tab:ab_study}
\end{table}

From Table~\ref{tab:ab_study}, we can find that \textbf{Basic} model performs better than \textbf{Basic-Mask} and \textbf{Basic-Aux}, which validates the usefulness of considering user utterances and speaker sentiments. \textbf{Basic+ST} perform better than \textbf{Basic} because speaker turn change provides helpful clues for utterance representation. Although TD is a simple classifier and its capability can be limited, \textbf{Basic+TD} still contributes stable and consistent performance improvements. The performance of \textbf{Basic+TD+ST} performs best among all these methods, which implies the effectiveness of different components.

\subsection{Influences of Adversarial Learning}
\begin{figure}
  \centering
      \includegraphics[width=4cm]{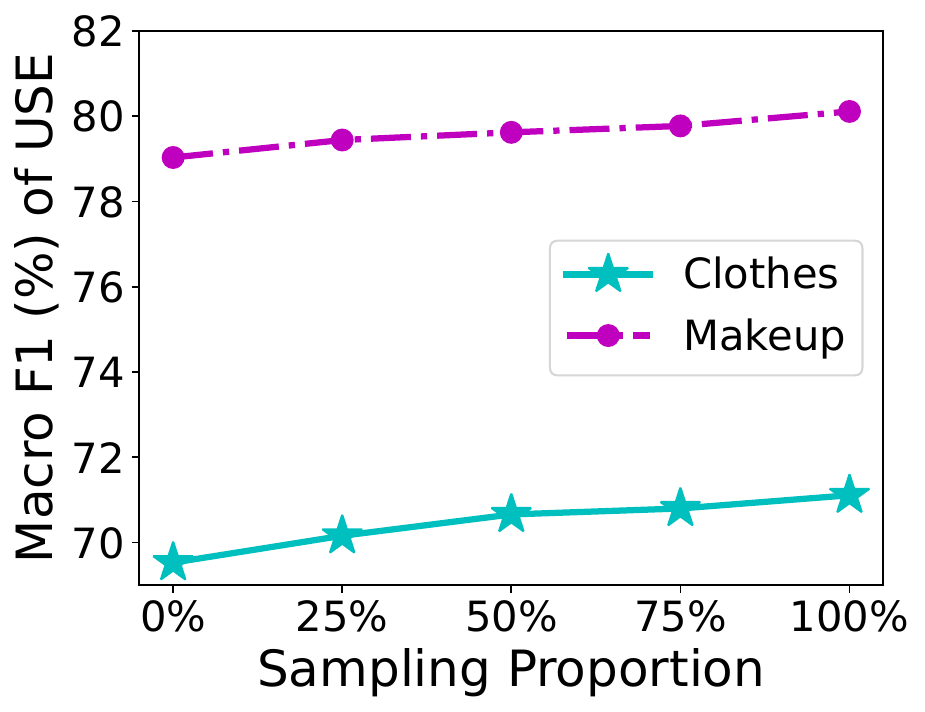}
      \includegraphics[width=4cm]{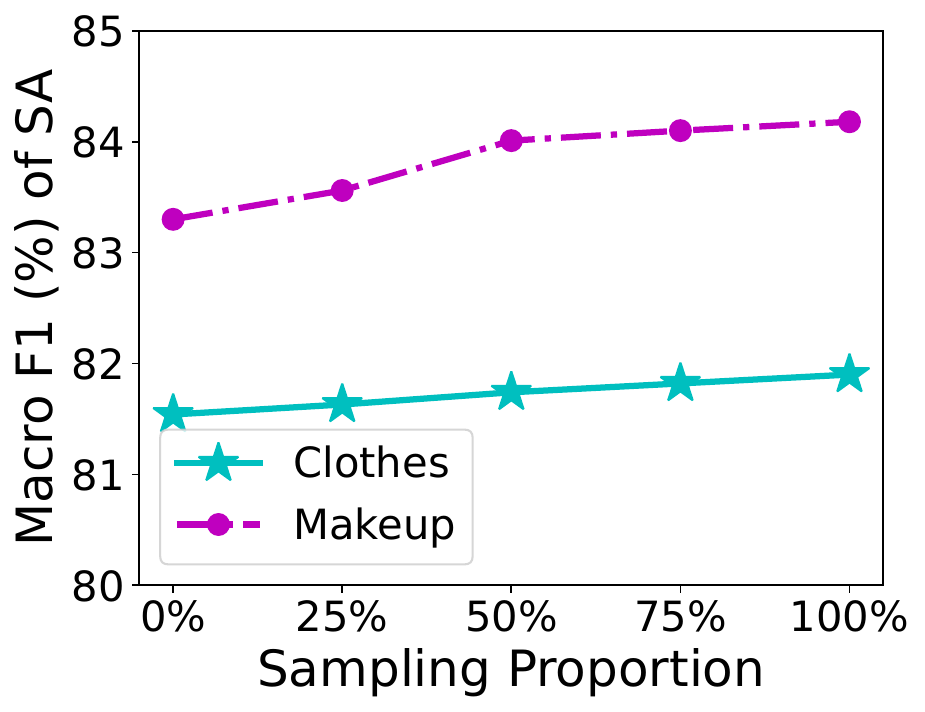}
  \caption{The influences of TD on USE and SA tasks.}
  \label{fig:td_influence}
\end{figure}

To investigate the influences of our multi-task adversarial learning strategy, we perform a in-depth study on \textit{Task Discriminator} (TD). Specifically, we randomly sample $x\%~~(x=0, 25, 50, 75, 100)$ training instances from the original training set, and further study the testing performance of USE and SA tasks, respectively. The experimental results are displayed in Figure~\ref{fig:td_influence}. 

From Figure~\ref{fig:td_influence}, we can find that our method performs better as the size of training data increases. Our method performs worst when $x=0$ because TD is not trained and it can not differentiate between USE and SA tasks. Our method performs quite well when $x\geq 75$ because TD achieves stable classification performance, which indicates that input features are more task-specific.

%To verify the above conclusion, we also study the influences of TD built from the sampled training set on the USE and SA tasks. The results are displayed in Figure~\ref{fig:td_influence} (middle and right ones). We can find that well trained TD will contribute to both USE and SA tasks. This again verifies the effectiveness of our adversarial learning strategy.

\subsection{Correlation between USE and SA}
\begin{figure}
  \centering    
  \includegraphics[width=6cm]{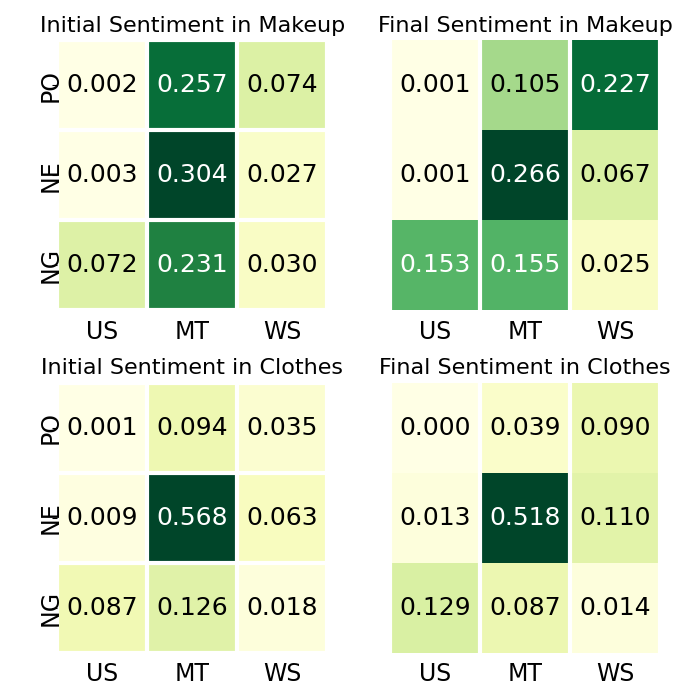}
\caption{The proportion of all the (initial/final user sentiment, user satisfaction) combinations over the datasets.}
\label{fig:consistency}
\end{figure}

To gain further insight into the correlation between USE and SA tasks, we enumerate all the (initial/final user sentiment, user satisfaction) combinations and calculate its proportion over the two datasets. Note that the initial/final user sentiment corresponds to the first/last user utterance in any dialogue. Figure~\ref{fig:consistency} displays the statistical results, from which we can find some interesting patterns: (i) Initial/final neutral sentiment will trigger any user satisfaction class, which is more intuitive on the \textit{Makeup} dataset. (ii) Initial negative sentiment is not effectively alleviated, but it gets worse at the end. Because the proportion of (US, NG) increases from 7.2\% to 15.3\% on \textit{Makeup} and 8.7\% to 12.9\% on \textit{Clothes}, respectively. (iii) Compared with initial sentiment, final sentiment plays more critical roles on deciding the user satisfaction. Specifically, we directly map the initial/final negative, neutral and positive sentiment to the corresponding user satisfaction, i.e., unsatisfied, met and satisfied, respectively. Given final sentiment, the total proportion of (US, NG), (MT, NE) and (WS, PO) combinations is significantly larger that under initial sentiments. All these verify that USE and SA are correlated tasks and can be learned together. 

Besides, we follow the heuristic \textit{sentiment-satisfaction} mapping method introduced above, and display the experimental results in Table~\ref{tab:consistency_cmp}. From the results, we can find that although SA and USE are highly pertinent tasks, sentiments can not be mapped to the user satisfaction directly, especially for \textit{met} satisfaction which has less connection with sentiments. This again verifies that SA task only provides limited helpful sentimental clues to assist with USE.
\begin{table}
  \centering
  \setlength\tabcolsep{1pt}
  \begin{tabular}{|l|c|c|c|c|}
  \hline
  % \rowcolor{gray!20}
  & \multicolumn{2}{c|}{\textbf{Clothes}} &\multicolumn{2}{c|}{\textbf{Makeup}}\\
    \cline{2-5}
    \multirow{-2}{*}{\textbf{Methods}} & \scriptsize{Macro F1} & \scriptsize{Accuracy} & \scriptsize{Macro F1} & \scriptsize{Accuracy}\\
    \hline
    Initial-User-Sentiment & 53.83  & 68.60  & 43.23 & 47.74 \\
    Final-User-Sentiment & 65.96  & 72.30  & 64.51 & 64.40 \\
    \textbf{STMAN} &\textbf{71.11} &\textbf{78.60} &\textbf{80.11} &\textbf{80.22}\\
  \hline
  \end{tabular}
  \caption{Initial/final user sentiments versus user satisfaction.}
  \label{tab:consistency_cmp}
\end{table}

\subsection{Case Study}
To better understand the usefulness of our model, we choose three example dialogues from \textit{Clothes} testing set and display the prediction results of Meta-LSTM, RSSN and STMAN in Table~\ref{tab:cases}. The original Chinese dialogues have been translated into English dialogues for understandability. We denote user as ``U" and service staff as ``S" to differentiate speakers.
\begin{table}
\newcommand{\tabincell}[2]{\begin{tabular}{c}#2\end{tabular}}
\centering
\small
  \begin{tabular}{|c|c|}
    % \rowcolor{gray!20}
    \hline
    \textbf{Exemplar Dialogues (Chinese $\rightarrow$ English)}\\
    \hline
    \tabincell{l}{\textbf{U:} Can you speed up logistics? Too slow! \textbf{(\scriptsize{NG,NG,NG,NG})}\\
    \textbf{S:} The logistics is running normally. \textbf{(\scriptsize{NE,NE,NE,NE})}\\-------\textit{Omit staff explanation}-------\\
    \textbf{U:} Try to help me speed up the progress. \textbf{(\scriptsize{NG,NG,NE,NE})}  \\\textbf{S:} Dear, I'll try my best to give feedback. \textbf{(\scriptsize{PO,PO,PO,PO})}\\\textbf{S:} Thanks for your support! Bye! \textbf{(\scriptsize{PO,PO,PO,PO})}\\
    \textbf{Meta-LSTM:}US ~\textbf{RSSN:}US ~\textbf{STMAN:}MT ~\textbf{Truth:}MT}\\
    \hline
   \tabincell{l}{\textbf{U:} Hello! \textbf{(\scriptsize{NE,NE,NE,NE})}\\
    \textbf{S:} What can I do for you, dear? \textbf{(\scriptsize{PO,PO,PO,PO})}\\
    \textbf{U:} What about the back of the pants? \\Do you have any pictures? \textbf{(\scriptsize{NE,NE,NE,NE})}  \\\textbf{S:} Dear, there is no real picture here. \textbf{(\scriptsize{PO,PO,PO,PO})}\\\textbf{S:} Dear, please check other people's reviews. \textbf{(\scriptsize{PO,PO,PO,PO})}\\
    \textbf{U:} If you don't know your product, how can you\\ introduce it to your customers? ? \textbf{(\scriptsize{NE,NE,NG,NG})}\\\textbf{S:}  Not every item has pictures. :):) \textbf{(\scriptsize{PO,PO,PO,PO})}\\
    \textbf{Meta-LSTM:}MT ~\textbf{RSSN:}MT ~\textbf{STMAN:}US ~\textbf{Truth:}US}\\
    \hline
   \tabincell{l}{\textbf{U:} I don't want ZTO Express. \textbf{(\scriptsize{NE,NE,NE,NE})}\\
    \textbf{U:} Are you there? \textbf{(\scriptsize{NE,NE,NE,NE})}\\
    \textbf{S:} Please wait a moment. \textbf{(\scriptsize{NE,NE,NE,NE})}\\-------\textit{No reply for a long time}-------\\\textbf{U:} ? ? ? \textbf{(\scriptsize{NE,NE,NE,NE})}\\
    \textbf{Meta-LSTM:}MT ~\textbf{RSSN:}MT ~\textbf{STMAN:}MT ~\textbf{Truth:}US}\\
    \hline
  \end{tabular}
  \caption{The predicted results of three example dialogues in \textit{Clothes}. The sentiment and satisfaction are displayed in Quadruple (Meta-LSTM, RSSN, STMAN, Truth).}
  \label{tab:cases}
\end{table}

For the 1$^{st}$ example, the user expresses negative sentiment in the beginning, but he is well comforted and expresses neutral sentiment in the end after the service staff explains the reason very sincerely (\textit{detail content is omitted}). However, Meta-LSTM and RSSN mispredict the 3$^{rd}$ utterance which is similar to the $1^{st}$ utterance in content. For the 2$^{nd}$ example, Meta-LSTM and RSSN can not predict the correct sentiment of 6$^{th}$ utterance which implies accusation. For the $3^{rd}$ example, all the three methods make wrong predictions because it lacks of useful information and its sentiment is dependent on time factor which is not considered in modeling. For the hard-to-solve case, we leave it for future study.

\section{Conclusion and Future Work}
In this paper, we propose a novel and extensible STMAN for joint USE and SA. Firstly, we introduce a basic model which adopts two hidden layers for each task: one is used to extract the common patterns via the shared parameters, and the other is used to capture task-specific features via the separate parameter set. Then, we enhance the basic model by introducing speak turn information to enhance the interaction layer. Finally, we introduce a multi-task adversarial strategy which makes inputs more task-specific via a task discriminator. Experiments conducted on two public service dialogue datasets indicate the effectiveness of our approach. 

In the future, we will investigate how to estimate user satisfaction in low-resource scenario.

\section{Acknowledgements}
We thank the anonymous reviewers for their valuable comments
and suggestions. This work is supported by National Natural Science Foundation of China (62106039) and National Key R\&D Program of China (2020YFC0832505).

\bibliography{aaai23}

\end{document}